\documentclass[letterpaper, 10 pt, conference]{ieeeconf}
\IEEEoverridecommandlockouts
\usepackage{amsmath}
\usepackage{amssymb}
\usepackage{graphicx}
\usepackage{url}
\usepackage{tikz}
\usetikzlibrary{positioning, arrows.meta, fit, backgrounds, calc}
\title{\LARGE \bf
From S3Q Theory to Implementation: Towards an Architecture for Machine Qualia
}
\author{Tetiana~Grinberg$^{1}$, Katrina~Schleisman$^{2}$, Patryk~Laurent$^{3}$, Bogdan~Udrea$^{4}$, Minda~Myers$^{5}$,
\\ Brian~Aufderheide$^{6}$,
Luis~El~Srouji$^{7}$, Doyle~Groves$^{8}$, and Kevin~Schmidt$^{9}$%
\thanks{$^{1}$Symbiokinetics Inc., {\tt\scriptsize tanya@symbiokinetics.com}}%
\thanks{$^{2}$Galois, Inc., {\tt\scriptsize katrina.schleisman@galois.com}}%
\thanks{$^{3}$Lighthill Technologies, Inc., {\tt\scriptsize patryk@lighthill.ai}}%
\thanks{$^{4}$VisSidus Technologies, Inc., {\tt\scriptsize bogdan.udrea@vissidus.com}}%
\thanks{$^{5}$Stanford University, {\tt\scriptsize minda@stanford.edu}}%
\thanks{$^{6}$Hampton University, {\tt\scriptsize brian.aufderheide@hamptonu.edu}}%
\thanks{$^{7}$UC Davis, {\tt\scriptsize lzelsrouji@ucdavis.edu}}%
\thanks{$^{8}$Indiana University, {\tt\scriptsize doygrove@iu.edu}}%
\thanks{$^{9}$Northwestern University, {\tt\scriptsize kevinschmidt2022@u.northwestern.edu}}%
\thanks{Code for toy implementations, along with additional material, will be made available at {\tt\scriptsize www.mind-guild.com}.}%
}
\begin{document}
\maketitle
\thispagestyle{empty}
\pagestyle{empty}
\begin{abstract}
A key challenge in machine consciousness research is translating theoretical models into computational-level implementations. In this paper, we address this challenge by proposing a five-layer implementation architecture for the S3Q (Simulated, Situated, Structurally Coherent) theory of consciousness~\cite{schmidt2021}. Rather than introducing novel formalisms, the architecture composes published computational primitives into a single pipeline. S3Q identifies three jointly necessary conditions for qualia: (1)~grounded sensorimotor \emph{situatedness}, (2)~internal \emph{simulation} via a world model, and (3)~\emph{structural coherence} between predictions and observations. No existing computational system implements all three simultaneously. We map each S3Q tenet to specific, compatible computational machinery and specify how these components interface within a single representation pipeline operating on continuous, differentiable, per-object slot vectors, together with a developmental bootstrap sequence and falsifiable predictions for the composed system that no subset of the architecture produces in isolation. The model suggests that a basic sense of ``self'' develops by linking actions to their outcomes, and that behavior falls into three patterns (hesitation, curiosity, or avoidance) depending on how unexpected an outcome is and whether it is experienced as positive or negative. Each prediction is individually falsifiable, providing the field with a testable framework to advance our understanding of machine consciousness.
\end{abstract}
\section{INTRODUCTION}
\subsection{The Problem}
The S3Q (Simulated, Situated, Structurally Coherent) framework~\cite{schmidt2021} identifies three conditions that must be jointly satisfied for machine consciousness: situatedness, simulation, and structural coherence. The theory specifies what these conditions mean at the functional level but does not specify how to implement them. The field has published components that address individual conditions in isolation: Slot Attention for object-centric parsing~\cite{locatello2020,kipf2022savi}, predictive coding for simulation~\cite{rao1999}, and precision-weighted prediction error for coherence monitoring. But no published system composes these into a single architecture satisfying all three S3Q tenets simultaneously, with defined interfaces between layers, a developmental bootstrap sequence, and falsifiable predictions that depend on the conjunction.
\subsection{The Claim}
We propose a five-layer architecture that implements the three S3Q tenets using published mathematical components. Each layer has a defined input/output interface, a specified training signal, and a developmental ordering. The contribution is not any individual component (all are published) but the composition: how they connect, what data flows between them, and what the conjunction predicts that no subset produces. The architecture specifies functional requirements, not a physical substrate; what matters is that the interfaces between layers are preserved. Multiple theoretical models frame consciousness as emergent: a global pattern arising from local interactions~\cite{krakauer2024}. The architecture makes two predictions that distinguish it from standard world-model agents and from disentangled~\cite{wang2024} representation learners: (a)~the agent's own body emerges as the slot with the highest mutual information with motor commands, and (b)~the agent exhibits qualitatively distinct behavioral responses to model violations depending on valence (hesitance, curiosity, avoidance). Both are individually falsifiable.
Consistent with the S3Q model, we define ``qualia'' in a strictly operational sense, a goal-situated, decomposable, structurally coherent representation, and provide initial testable claims about phenomenal consciousness. Whether an agent with these representational properties has subjective experience is a question for future work.
\subsection{The S3Q Framework}
The S3Q framework~\cite{schmidt2021} identifies three tenets that must be satisfied for a system to exhibit machine consciousness:
\textbf{Situated}: All aspects of the representation are defined by relationships and interactions that take place in a grounded and dynamic world.
\textbf{Simulated}: The representation is an internally generated sensorimotor world model.
\textbf{Structurally Coherent}: The representation captures enough information about the agent's reality to facilitate grounding through environmental interaction. It represents a balance between overlearned rigid habits and flexible cognition.
\subsection{Definition of Qualia}
\label{sec:qualia}
We define a quale as a processed feature that the agent's goals have made relevant and that its internal world model can use directly. For example, a typical person admiring a landscape would not notice reflections on the surfaces of objects, but a photographer would, because their training makes such features relevant to their goals. Section~\ref{sec:arch} gives the architectural counterparts: slots (Layers~1 and~2) provide separability, precision modulation (Layer~5) provides goal-sensitivity, and the forward model~$T$ provides the reference relation.
\section{ARCHITECTURAL OVERVIEW: FIVE FUNCTIONAL LAYERS}
\label{sec:arch}
\begin{figure*}[t]
    \centering
    \includegraphics[width=\linewidth]{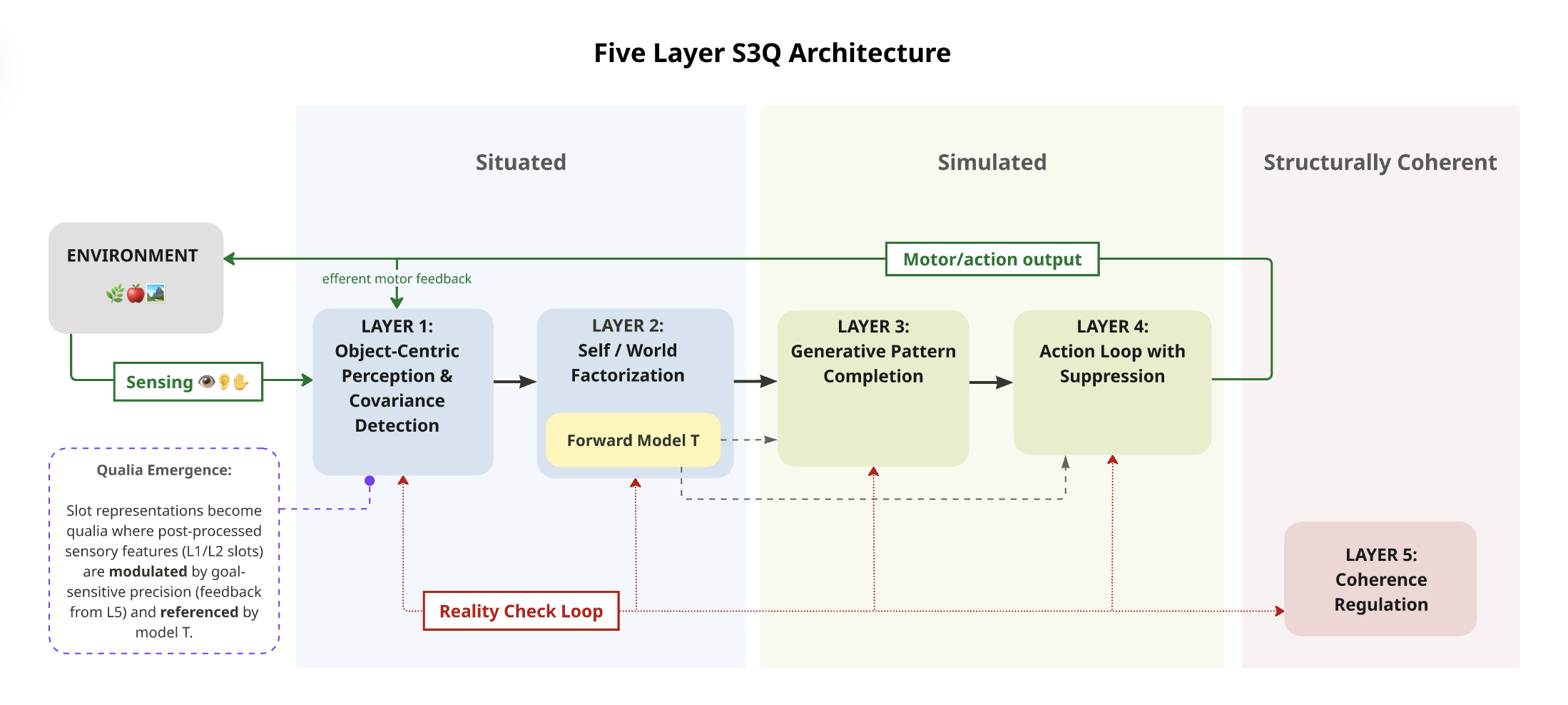}
    \caption{Five-layer S3Q architecture. Layers are colored by tenet (blue: Situated, green: Simulated, pink: Structurally Coherent). Solid arrows show the action/data loop; dashed arrows show reuse of the forward model~$T$; dotted arrows show coherence regulation (functionally attributed to Layer~5).}
    \label{fig:placeholder}
\end{figure*}

We decompose the three S3Q tenets into five functional layers, each with specific mathematical requirements. The layers are ordered developmentally: earlier layers bootstrap later ones.
\begin{itemize}
    \item \textbf{Situated} $\rightarrow$ L1: Object-centric perception \& covariance detection; L2: Self/world factorization, forward model~$T$. \emph{Core question:} How does the agent parse the sensory stream into objects, characterize dynamics w.r.t.\ its own actions, and identify its own body?
    \item \textbf{Simulated} $\rightarrow$ L3: Pattern completion; L4: Action loop with suppression. \emph{Core question:} How does the agent ``imagine'' unobserved states and decide whether to execute or suppress motor output?
    \item \textbf{Structurally Coherent} $\rightarrow$ L5: Coherence regulation. \emph{Core question:} How does the agent detect when its internal model has gone wrong, and what does it do about it?
\end{itemize}
A \textbf{slot} is a fixed-length vector $\mathbf{s}_t^i \in \mathbb{R}^{d}$ representing a single object's features (position, shape, color, motion) as a single unit, maintained separately from all other objects throughout the pipeline. The superscript $i \in \{1, \dots, K\}$ indexes the slot and the subscript indexes time; every slot shares the common dimension~$d$. All five layers converge on a shared representation: continuous per-object slot vectors. Layer~1 produces the slots and characterizes each slot's dynamics. Layer~2 uses those characterizations to label each slot as self or world and trains the single forward model~$T$ that the entire system uses. Layers~3 through~5 operate on slots using~$T$. Representation learning in Layers~1 through~3 is unsupervised or self-supervised (the agent's own motor stream provides conditioning signals, but no external reward or labels are required). Behavioral policy selection in Layer~5 additionally depends on a valence signal (Section~\ref{sec:coherence}). Fig.~\ref{fig:placeholder} illustrates the full architecture.
\textbf{Developmental bootstrap.} Layer~1 converges first in two stages. Stage~1: an object-centric encoder (e.g., Slot Attention~\cite{locatello2020} or SAVi~\cite{kipf2022savi}) is trained to convergence on raw sensory input, producing stable per-slot trajectories. Stage~2: per-slot slow-feature analysis (SFA) and mutual information (MI) estimation run on the stabilized trajectories to identify action-contingent slots. Layer~2 (self/world labeling and forward model~$T$) is trained once Layer~1's MI estimates stabilize. Layer~4 begins action evaluation using~$T$ as soon as it is available. Hopfield patterns accumulate during normal operation. Once the suppression gate of Layer~4 produces reliable motor modulation, the system switches to imagination mode. Layer~5 is passive until~$T$ and the suppression gate produce stable predictions.
\section{TENET 1: SITUATED (LAYERS 1 AND 2)}
\subsection{Layer 1: Object-Centric Perception and Covariance Detection}
S3Q emphasizes \emph{situatedness}: cognition occurs on-the-fly by a dynamic agent in a dynamic world. The architecture addresses this by requiring two capabilities: (1)~parsing the continuous sensory stream into discrete objects, and (2)~characterizing each object's dynamics relative to the agent's own actions.
\textbf{Object-centric encoding.} Layer~1 requires an encoder that decomposes sensory input into $K$ discrete slot vectors of common dimension~$d$ via competitive attention: features assigned to one slot are unavailable to others, preventing any single slot from absorbing the entire scene. Each slot captures one object's shape, color, position, and motion. For temporally extended input (e.g., a frame-based video stream), the agent must maintain slot-to-object identity across frames without an external matching rule. Slot Attention~\cite{locatello2020} and SAVi~\cite{kipf2022savi} satisfy these requirements; more recent methods (e.g., VideoSAUR~\cite{zadaianchuk2023}) could serve as drop-in replacements. The architecture requires the recurrent-slot-binding pattern, not any specific method. For multimodal input, each modality is encoded by its own head, and the resulting feature tokens are concatenated into a single input set for slot decomposition.
\textbf{Periodicity bias and Slow Feature Analysis.} Stereotypic reflexive actions (saccades, head-bobbing, limb cycling, rhythmic probing) are the cheapest way to generate sensory contrast. The architecture requires a mechanism to extract stable, slowly varying features from each slot's trajectory, distinguishing real structure from noise. Slow Feature Analysis (SFA)~\cite{wiskott2002} satisfies this; its incremental version (IncSFA~\cite{kompella2012}) handles non-stationary environments with biologically plausible update rules. Luciw et al.~\cite{luciw2013} combine IncSFA with curiosity-driven exploration, demonstrating autonomous development of multiple invariant representations.
\textbf{Contrastive mutual information estimation} complements per-slot SFA by estimating $I(A; \mathbf{s}^i)$ between the agent's actions and each slot's feature vector, where $A$ denotes the action random variable and $\mathbf{u}_t$ (Section~\ref{sec:forwardmodel}) denotes its realization at time~$t$. MINE~\cite{belghazi2018} and InfoNCE~\cite{vanoord2019} both suffice for the ranking task (identifying which slots have highest MI with actions). Together, per-slot SFA and MI estimation identify slots whose trajectories are both stable (slow) and controllable (high MI with actions). These are the candidates for ``self'' in Layer~2.
\subsection{Layer 2: Self/World Factorization}
\label{sec:selfworld}
\textbf{Self/world labeling.} Slots with $I(A; \mathbf{s}^i)$ above a threshold $\theta_I$ are attributed to self (typically a single slot: the agent's own body); the remainder are attributed to world. The threshold is set relative to the MI distribution across slots rather than as an absolute value, since absolute MI scales with sensor resolution and action dimensionality. When more than one slot exceeds it, all are retained as self, which is the correct behavior for agents with articulated bodies parsed into multiple slots. We predict that the agent's own body will emerge as the slot with highest MI with its motor commands. This is a key empirical claim, not a guaranteed consequence of the architecture.
Two caveats bound this claim. First, in environments where the body is the only action-contingent object, the prediction holds by construction rather than as a finding about the architecture. The discriminating test requires an environment containing a second action-contingent object, such as a continuously controlled tool or a mirror image of the agent. If the architecture labels the tool as self, that result is informative either way, and it separates a genuine self/world factorization from the trivial case where only one thing moves with the motors. Second, for agents whose body lies largely outside their own sensory field, such as a purely visual saccading agent, self-motion produces correlated flow across all slots rather than a single high-MI slot. The architecture as specified assumes the body is parsed as an object in the sensory field.
\textbf{Encapsulation.} Self/world factorization is only meaningful if slot contents stay separated, which two mechanisms downstream and upstream of Layer~2 jointly guarantee. Slot Attention (Layer~1) enforces encapsulation of features within object boundaries in the encoder. The per-slot dynamics model~$T$ (Section~\ref{sec:forwardmodel}) prevents feature smearing during forward simulation: each slot's transitions depend only on that slot's own state. These are complementary mechanisms.
\textbf{Static composition (nouns).} An agent must learn to deconvolve stable features (e.g., shape, color) from features specific to a given exposure context (e.g., a cast shadow, a specular reflection). Which of these count as incidental is itself goal-dependent: the same specular reflection is discardable for an agent navigating a room and diagnostic for one estimating surface material (Section~\ref{sec:qualia}). Features become qualia only once situated within the agent's goal hierarchy.
\subsection{The Forward Model ($T$)}
\label{sec:forwardmodel}
\textbf{Factored dynamics model.} We propose a per-slot transition function
\begin{equation}
\label{eq:T}
T: \big(\mathbf{s}_t^i,\, \mathbf{u}_t,\, \mathbf{c}^i\big) \mapsto \big(\Delta \hat{\mathbf{s}}_t^i,\, (\sigma_{t+1}^i)^2\big), \quad
\hat{\mathbf{s}}_{t+1}^i = \mathbf{s}_t^i + \Delta \hat{\mathbf{s}}_t^i,
\end{equation}
where $\mathbf{s}_t^i$ is the current state of slot~$i$, $\mathbf{u}_t \in \mathbb{R}^{d_u}$ is the ``action'' (motor command if $\ell^i = \text{self}$, inferred external cause if $\ell^i = \text{world}$), and $\ell^i \in \{\text{self}, \text{world}\}$ is the per-slot source label from Layer~2 with learned embedding $\mathbf{c}^i \in \mathbb{R}^{d_c}$. Predicted quantities carry the index of the timestep they describe, so $\hat{\mathbf{s}}_{t+1}^i$ and $(\sigma_{t+1}^i)^2$ both refer to the prediction for time~$t+1$ formed at time~$t$. Source attribution is refined iteratively as the dynamics model improves, in an Expectation-Maximization (EM)-like alternation between source inference and dynamics learning. Consistent with the efference-copy tradition~\cite{wolpert1995,crapse2008}, the source can also be read off the residual of an unconditional forward model. Convergence of this procedure is a target for empirical validation. $T$~is self-supervised: trained on $(\mathbf{s}_t^i, \mathbf{u}_t, \mathbf{s}_{t+1}^i)$ triples from the agent's own motor stream.
The functional form of~$T$ is identical for self-caused and externally-caused transitions: a pushed object follows the same trajectory regardless of who initiated the push. The source label enters only through the small learned embedding $\mathbf{c}^i$, which modulates~$T$'s output, allowing sufficient expressivity without duplicating the model. $T$ is introduced here because Layer~2 trains it, but $T$ is a single shared component rather than a layer: Layers~3 through~5 reuse it without retraining.
\textbf{One forward model, varying input completeness.} $T$~is used across multiple layers with inputs of varying completeness. In perception, the input is the actual sensory observation. In imagination, the input is $T$'s own previous output plus whatever proprioceptive signal leaks through the suppression gate (Section~\ref{sec:suppression}). This forward-model-plus-controller pattern traces to Schmidhuber's early work on differentiable world models~\cite{schmidhuber1990} and Ha \& Schmidhuber 2018~\cite{ha2018}. The specific contribution here is to make~$T$ slot-structured, train it to output both the predicted next state and a per-state variance estimate, and connect it to a precision-weighted coherence monitor. The variance estimate matters because prediction uncertainty is not uniform: some states are inherently harder to predict (a ball near a table edge vs.\ resting on the floor), and the model must learn where its own predictions are reliable.
The architecture requires~$T$ to be a per-slot transition function that preserves slot identity and produces state deltas. C-SWM~\cite{kipf2020cswm} (graph neural network with contrastive objective) is well-matched to short-horizon predictions. For long-horizon imagination rollouts, a SlotFormer-style~\cite{wu2023} autoregressive Transformer over slot tokens is more appropriate. Either way~$T$ is trained once and reused.
\textbf{Precision estimation via heteroscedastic regression.} $T$~outputs two quantities per slot per timestep: a predicted next state (the mean) and a predicted variance~$(\sigma_{t+1}^i)^2$ (the confidence), trained jointly via negative log-likelihood (NLL)~\cite{kendall2017}. Prediction error is defined per slot and aggregated across slots:
\begin{equation}
\label{eq:errchain}
e_t^i = \big\| \hat{\mathbf{s}}_t^i - \mathbf{s}_t^i \big\|_2, \quad
\epsilon_t^i = e_t^i / \sigma_t^i, \quad
\epsilon_t = \max_i \epsilon_t^i,
\end{equation}
where $\| \cdot \|_2$ is the Euclidean norm, $\epsilon_t^i$ is the per-slot precision-weighted prediction error, and the aggregate~$\epsilon_t$ is the error of the worst-offending slot. Section~\ref{sec:coherence} uses~$\epsilon_t$ to select a behavioral response. This mirrors Friston's framework where precision is encoded by synaptic gain of prediction-error neurons~\cite{friston2005,bastos2012}, and the brain optimizes predictions and their precision simultaneously~\cite{friston2005,friston2017}. A known risk is variance overestimation; mitigations include the $\beta$-NLL loss~\cite{seitzer2022}, whose weighting exponent is the only role of the glyph~$\beta$ in this paper, and minimum-precision constraints.
\textbf{Pattern completion via Modern Hopfield Networks.} Modern Hopfield Networks provide associative memory via energy minimization, with polynomial storage capacity in the dense formulation~\cite{krotov2016} and exponential capacity in the continuous formulation~\cite{ramsauer2020}. Ramsauer et al.~\cite{ramsauer2020} prove the modern Hopfield update rule is equivalent to transformer attention, making it architecturally compatible with Slot Attention. A single Hopfield network stores concatenated slot vectors as patterns, enabling cross-slot pattern completion. Its contribution scales with input incompleteness: full observations barely change under Hopfield dynamics, whereas sparse imagined representations change substantially, so imagination serves as a \emph{seed} to fill out missing information in the world model. For example, hearing barking and glimpsing a wagging tail behind a fence would be enough to complete the pattern and infer a dog.
\section{TENET 2: SIMULATED (LAYERS 3 AND 4)}
\subsection{Layer 3: Pattern Completion (Imagination from Sparse Seeds)}
S3Q posits that all experience is a form of simulation: what we experience is our world model, not the actual world. Layers~3 and~4 address top-down cognition, leveraging past experience to fill in gaps when novel but similar information is encountered.
Given a partial or noisy representation (a sparse seed), the agent must complete it to a full representation consistent with learned structure. This serves three functions: (a)~fast recognition in real time, (b)~dreaming (body signals during sleep act as sparse seeds), (c)~filling in the forward model's output during imagination.
\textbf{Dynamic composition (verbs).} Agents must learn ``operators'' (e.g., bouncing, falling, walking) that can be applied to any suitable object noun. A verb is a learned transition kernel~$\Phi: \mathbb{R}^{d} \rightarrow \mathbb{R}^{d}$; a noun is a slot feature vector~$\mathbf{z} \equiv \mathbf{s}_t^i$; composition is $\Phi(\mathbf{z})$, yielding the predicted next-state slot vector.
\textbf{The Noun-Verb Composition Test.} The architecture predicts that an agent can simulate a novel configuration (e.g., a Blue Ball Bouncing) never seen before by applying a learned transition kernel to a slot in a novel combination. C-SWM~\cite{kipf2020cswm} demonstrates combinatorial generalization to novel object configurations. The formal backbone is the Markov Decision Process (MDP) homomorphism framework~\cite{vanderpol2020}: when the contrastive loss reaches zero, the learned dynamics are an exact homomorphism of the original MDP, guaranteeing that transitions in latent space commute with transitions in observation space.
\subsection{Layer 4: Action Loop with Continuous Suppression}
\label{sec:suppression}
Layer~4 implements motor suppression so that internal representations can be activated for simulation without automatically invoking motor responses, a capability needed for imagination and dreaming~\cite{dresler2011,guillot2012}. In many brain systems, top-down feedback connections outnumber bottom-up ones~\cite{briggs2020} and serve as cognitive control. The agent continuously (a)~proposes candidate actions, (b)~projects each through~$T$, (c)~evaluates the projected outcome against current objectives, and (d)~modulates motor output via a continuous suppression gate $g_t \in [0, 1]$.
\textbf{Motor output is always being generated.} Imagination requires suppression, not activation. The default state is action. This is consistent with basal ganglia function: tonic inhibition from the internal segment of the globus pallidus (GPi) and the substantia nigra pars reticulata (SNr) must be released for action to occur~\cite{mink1996}.
\textbf{The suppression gate.} Functionally, the gate integrates two signals: (1)~an error-driven signal that increases suppression when the world model is surprised (high $\epsilon_t$, Eq.~\ref{eq:errchain}), causing the agent to pause before acting on uncertain predictions; and (2)~a context-driven signal $g_\text{ctx}$ set by behavioral state (high during imagination/sleep, low during exploration). One possible realization maps these through a widely used sigmoid:
\begin{equation}
\label{eq:gate}
\begin{gathered}
g_\text{err} = \text{sigmoid}\big(\kappa\, (\epsilon_t - \theta_g)\big), \quad
g_t = \max(g_\text{err},\, g_\text{ctx}), \\
\mathbf{m}_t = (1 - g_t)\, \mathbf{u}_t,
\end{gathered}
\end{equation}
where $\kappa > 0$ sets the steepness of the error gate, $\theta_g$ is its midpoint (the surprise level at which the gate is half open), and $\mathbf{m}_t$ is the motor output delivered to the actuators. Reflex pathways override both gates by forcing $g_t = 0$.
\textbf{Leaky suppression} ($g_t \approx 0.9$) produces subthreshold motor output during sleep, generating proprioceptive seeds for pattern completion. We hypothesize that dreams are what the pattern completion system produces when seeded by leaked motor output.
\textbf{Action evaluation via $T$.} For each candidate action~$\mathbf{a}_j$, $j \in \{1, \dots, J\}$, the agent computes $T(\mathbf{s}_t^i, \mathbf{a}_j, \mathbf{c}^i)$ with $\ell^i = \text{self}$, scores the projected state against current objectives as $q_j$, and selects $j^\ast$ by softmax with temperature~$\tau$; the executed action is $\mathbf{u}_t = \mathbf{a}_{j^\ast}$. Successor Features~\cite{barreto2017,dayan1993} provide an optional alternative when transfer across multiple goals is required.
\section{TENET 3: STRUCTURALLY COHERENT (LAYER 5)}
\label{sec:coherence}
Structural coherence in S3Q is a ``just right'' constraint: the world model must be close enough to reality to support adaptive action, yet abstract enough to support generalization under uncertainty. The coherence computation compares predictions to observations, weighted by confidence. We describe this as a distinct functional layer for expository clarity, but the architecture does not require a dedicated monitoring subsystem. The computation could be realized by a separate module, locally within each slot's own processing (as in O'Reilly's eXtended Contrastive Attractor Learning (XCAL) rule, where temporal difference computes prediction error within a single unit without separate error-encoding neurons), or as a multi-scale combination of both. The architecture requires only that the quantities below are computed; it does not prescribe where.
\textbf{The key quantities}, with the error chain as defined in Eq.~\ref{eq:errchain}:
\begin{itemize}
\item $e_t^i = \| \hat{\mathbf{s}}_t^i - \mathbf{s}_t^i \|_2$ \quad (per-slot prediction error, Euclidean norm)
\item $(\sigma_t^i)^2$ = predicted variance for slot~$i$ from $T$'s heteroscedastic head
\item $\epsilon_t^i = e_t^i / \sigma_t^i$ \quad (per-slot precision-weighted prediction error)
\item $\epsilon_t = \max_i(\epsilon_t^i)$ \quad (worst-offending slot drives behavioral response)
\item $v_t = \hat{r}_t$ = predicted valence, the output of a learned head trained to predict the reward signal~$r_t$; negative values are aversive, positive appetitive
\end{itemize}
\textbf{Three-zone response based on $\epsilon_t$ and $v_t$:}
The three-zone structure is a distinguishing prediction of this framework. Precision-weighted prediction error measures surprise, not badness. Partitioning ``large surprise'' into distinct behavioral modes requires valence. Mode selection is a stateless function of $(\epsilon_t, v_t)$. Two learnable thresholds govern the rule: $\theta_\epsilon$ (on surprise) and $\theta_v$ (on aversive valence):
\begin{enumerate}
\item \textbf{Hesitance} (moderate $\epsilon_t$): Increase $g_t$. The agent pauses motor output while continuing to process internally. This is the fast response and does not require valence.
\item \textbf{Curiosity} (large $\epsilon_t$, neutral/appetitive $v_t$): Switch to exploration. Generate contrast-producing actions near the surprising stimulus. Decrease $g_t$. This corresponds to the compression-progress / intrinsic-motivation tradition~\cite{schmidhuber1991,schmidhuber2010}.
\item \textbf{Avoidance} (large $\epsilon_t$, aversive $v_t$): Switch to withdrawal. Increase distance from the surprising stimulus.
\end{enumerate}
This piecewise structure is a simplification of Expected Free Energy in active inference~\cite{friston2017}, where epistemic and pragmatic value jointly determine policy selection.
\textbf{Goal-sensitive precision modulation.} Goals shape which features become qualia via precision-weighted modulation of Layer~1 representations. Goal-relevant slots receive higher precision weights~$\alpha_i$, amplifying their prediction errors and driving faster representation learning. This is the Fristonian account of attention as precision optimization~\cite{friston2005,friston2017,feldman2010}. The result: two agents with identical sensory exposure but different goal structures develop different decompositions. The same precision weights modulate $T$'s per-slot learning rate~$\eta_i$, for example proportionally as $\eta_i = \eta_0 \alpha_i$ with $\eta_0$ the base learning rate of~$T$: high-precision slots generate larger gradient signals, so $T$ improves faster on goal-relevant objects. Precision also governs the balance between pattern completion and sensory input in Layer~3: when coherence is low, the Hopfield contribution is attenuated by a mixing coefficient $\lambda_t \in [0,1]$ in favor of direct observation.
\textbf{Coherence index.} The general coherence index~$C_t$ is the inverse of the mean precision-weighted error across all slots, aggregated over a recent window of $W$ timesteps:
\begin{equation}
\label{eq:coherence}
C_t = \Big[ \tfrac{1}{W} \textstyle\sum_{t'=t-W+1}^{t} \tfrac{1}{K} \textstyle\sum_{i=1}^{K} \epsilon_{t'}^i \Big]^{-1}.
\end{equation}
Linear probing~\cite{alain2016} provides an experimental diagnostic: object identity (``nouns'') must remain decodable throughout dynamic simulation (``verbs'').
\section{DISTINGUISHING CLAIMS AND FALSIFIABILITY}
\label{sec:falsify}
\subsection{What This Framework Claims That Alternatives Do Not}
Each component is published. The contribution is the composition. A Slot Attention system alone produces disentangled representations but does not ground them in actions or regulate them via prediction error. A Dreamer-style world model~\cite{hafner2020} simulates but does not differentiate behavioral responses by valence. Three claims depend on the conjunction:
\textbf{Claim 1: Situatedness determines which features become qualia.} The agent's goal hierarchy selectively determines which features are decomposed at high fidelity. Two agents with identical sensory exposure but different goals should develop different decompositions.
\textbf{Claim 2: Active exploration produces qualitatively different representations than passive observation.} The self slot, or slots for an articulated body, should emerge as those with the highest MI with the motor command channel, a signal that exists only under active exploration. A passive observer cannot identify the self slots through this mechanism.
\textbf{Claim 3: Coherence regulation produces adaptive behavioral modes.} When precision-weighted prediction error is high, the agent exhibits one of three qualitatively distinct responses depending on valence. A standard world-model agent does not exhibit hesitance or mode-switching between curiosity and avoidance.
\subsection{Falsification Criteria}
The framework is falsified if any of the following are obtained:
\begin{enumerate}
\item \textbf{No situatedness effect.} Attenuating or ablating Layers~2 and~5 produces equivalent representation quality.
\item \textbf{No active exploration advantage.} Passive observation produces equivalent identification of the self slots.
\item \textbf{No behavioral mode differentiation.} The response to all model violations is uniform regardless of valence.
\item \textbf{Feature smearing under composition.} Object identity collapses during imagination (noun-verb composition test fails).
\item \textbf{No imagination advantage.} Compositional imagination does not outperform a system requiring every specific combination during training.
\end{enumerate}
\subsection{Environment Requirements}
Any test environment must provide: (1)~multiple discrete objects with independently varying features and dynamics, at least two of them action-contingent, so that self-slot identification is discriminating rather than true by construction (Section~\ref{sec:selfworld}), (2)~action-dependent observability, (3)~sufficient continuous physics for prediction error signals, and (4)~goal flexibility so the same objects can be relevant under one goal and irrelevant under another. Environments meeting these criteria include ThreeDWorld~\cite{gan2020} and MineDojo~\cite{fan2022}.
\section{DISCUSSION}
\subsection{Relation to Existing Work}
The framework builds on but is not reducible to disentangled representation learning~\cite{wang2024}. The additional requirements (goal-sensitive feature selection, active sensorimotor grounding, coherence regulation with valence-dependent behavioral modes) distinguish it from a capable autoencoder.
The architecture draws substantially on Active Inference~\cite{friston2005,friston2009}: precision-weighted prediction error, precision-as-attention, and the piecewise behavioral response as a simplification of Expected Free Energy~\cite{friston2017}. However, it treats the suppression gate and mode-selection rule as engineered components rather than deriving them from a single variational objective, and does not adopt the full Free Energy Principle apparatus.
\subsection{Scope Limitations and Open Questions}
Theory of mind, multi-agent interaction, action repertoire learning, convergence of the source-attribution bootstrap, behavior of cross-slot Hopfield pattern completion under high occlusion, and neuromodulatory contextual calibration remain open questions for future work.
\section{CONCLUSION}
The contribution of this work is the composition: how published components connect via a shared slot-vector representation, what the conjunction predicts that no subset produces, and the falsification criteria (Section~\ref{sec:falsify}) that ensure either outcome advances the field. The architecture's modularity also suggests applications beyond consciousness research. It complements existing parametric tools in computational psychiatry~\cite{adams2013,mathys2011} by enabling \emph{structural} perturbations: disabling the suppression gate, distorting Layer~5 precision weights, or disrupting Layer~2 self/world factorization~\cite{friston2005,bastos2012}. For embodied AI, the same modularity provides explicit attachment points for safety constraints and failure diagnostics. Whether the architecture succeeds or falls short, the falsification criteria ensure that either outcome is informative.
\section*{ACKNOWLEDGMENT}
The authors thank Ted Selker for his valuable feedback on the manuscript, and the Mind Guild, of which the authors are members, for providing our platform to engineer the mind. We also thank the ACT3 group at the Air Force Research Laboratory, both for valuable discussion and for much of the inspiration behind this work. This work was not sponsored or funded by any organization; the acknowledgments above reflect intellectual contributions only.
\addtolength{\textheight}{-7cm}

\end{document}